\documentclass{article}

\usepackage[preprint]{neurips_2026}

\usepackage[utf8]{inputenc}
\usepackage[T1]{fontenc}
\usepackage{hyperref}
\usepackage{url}
\usepackage{booktabs}
\usepackage{amsfonts}
\usepackage{amsmath}
\usepackage{amssymb}
\usepackage{nicefrac}
\usepackage{microtype}
\usepackage{xcolor}
\usepackage{graphicx}
\usepackage{subcaption}
\usepackage{multirow}
\usepackage{makecell}
\usepackage{xspace}
\usepackage{float}

\newcommand{\cycleknn}{\textsc{cycle-kNN}\xspace}
\newcommand{\ours}{WRH\xspace}

\title{The Wittgensteinian Representation Hypothesis: \\ Is Language the Attractor of Multimodal Convergence?}

\author{%
  \textbf{Zhaoyang Zhang}$^{1,4}$\thanks{Work done during the internship at Ant Group.}\enspace
  \textbf{Run Shao}$^{1}$\enspace
  \textbf{Dongyue Wu}$^{2}$\enspace
  \textbf{Jiajie Teng}$^{3}$ \\
  \textbf{Chao Tao}$^{1}$\enspace
  \textbf{Jingdong Chen}$^{4}$\thanks{Corresponding authors.}\enspace
  \textbf{Haifeng Li}$^{1}$\footnotemark[2] \\[1ex]
  {\small $^{1}$Central South University \quad $^{2}$Huazhong University of Science and Technology} \\
  {\small $^{3}$Shanghai Jiao Tong University \quad $^{4}$Ant Group}
}

\begin{document}

\maketitle

\begin{abstract}
Understanding why independently trained neural networks from different modalities converge toward shared representations, and where this convergence leads, remains an open question in representation learning. All existing evidence relies on symmetric similarity measures, which can detect convergence but are structurally blind to its direction. We introduce \textbf{directional convergence analysis} using \cycleknn, an asymmetric alignment measure, applied across dozens of independently trained unimodal models spanning point clouds, vision, and language. We uncover a consistent directional asymmetry: non-language modalities move toward the neighborhood structure of language significantly more than the reverse, and this pattern holds across all model families and scales---yet is entirely invisible to symmetric measures. Mechanistic analysis traces the directionality to feature density asymmetry, whereby language representations occupy the most compact regions of representational space. The Information Bottleneck framework provides a principled interpretation: optimization under compression drives representations toward discrete, compositional structures characteristic of language. We formalize this as the \textbf{Wittgensteinian Representation Hypothesis}: the semantic structure of language is the asymptotic attractor of multimodal representation convergence.
\end{abstract}

\section{Introduction}
\label{sec:intro}

Understanding why representations from different modalities tend to converge, and identifying the endpoint of such convergence, is a fundamental question in multimodal representation learning. The Platonic Representation Hypothesis (PRH)~\citep{huh2024platonic} posits that independently trained models converge toward shared representational structures, an observation reinforced by theoretical work~\citep{lu2025indra, ziyin2025neural, tasker2026une} and empirical demonstrations of cross-modal alignment without paired data~\citep{schnaus2025blind, maniparambil2025frozen}.

However, \textbf{all empirical evidence relies on symmetric similarity measures}---CKA~\citep{kornblith2019similarity}, mutual kNN~\citep{huh2024platonic}, RSA~\citep{kriegeskorte2008representational}---which by construction satisfy $s(\mathbf{X}, \mathbf{Y}) = s(\mathbf{Y}, \mathbf{X})$. No prior work employs an asymmetric measure to investigate convergence \emph{direction}~\citep{klabunde2025similarity}. Like a ball rolling into a valley, if representation convergence possesses directional structure, symmetric measures are fundamentally blind to it.

\begin{figure}[t]
  \centering
  \includegraphics[width=\textwidth]{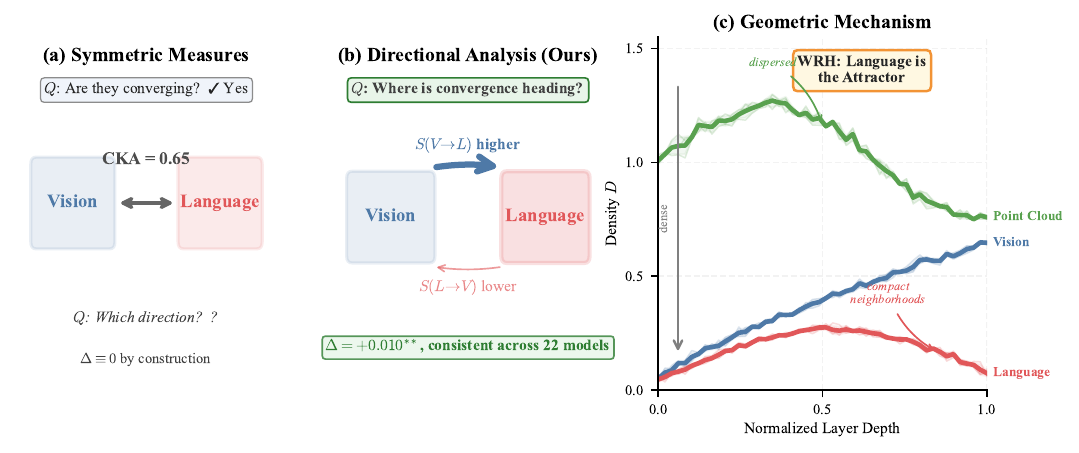}
  \caption{\textbf{Overview of directional convergence analysis.} (a)~Symmetric measures (e.g., CKA) detect convergence but cannot reveal its direction. (b)~\cycleknn is inherently asymmetric; analyzing both directions reveals a consistent directional bias: vision representations approach language more than the reverse ($\Delta = +0.010$, $p < 0.05$, across all 22 model pairs). (c)~Three modalities span an abstraction hierarchy; representational compactness ($D$) increases toward language. \ours posits that language is the attractor of multimodal convergence.}
  \label{fig:overview}
\end{figure}

We fill this gap with \textbf{directional convergence analysis} (Figure~\ref{fig:overview}). Our key insight is that \cycleknn---a known alignment measure whose inherent asymmetry has been consistently overlooked~\citep{huh2024platonic}---naturally encodes convergence direction. We systematically evaluate 58 independently trained unimodal models spanning three modalities along an \textbf{abstraction hierarchy}: 3D point clouds $\rightarrow$ 2D vision $\rightarrow$ language. Our core findings:

\begin{itemize}
    \item \textbf{Convergence has direction.} \cycleknn reveals consistent asymmetry: vision and point cloud representations approach language significantly more than the reverse---invisible to symmetric measures on the same pairs.
    \item \textbf{Language shows maximal intra-modality consensus}, consistent with proximity to the attractor.
    \item \textbf{Directionality is scale-invariant.} $\Delta > 0$ toward language holds for every vision model (22/22), every point cloud model (7/7), and nearly every language model across ten families spanning four orders of magnitude.
    \item \textbf{Feature density asymmetry as geometric mechanism.} Language representations are among the most compact; the Information Bottleneck framework provides a principled interpretation.
\end{itemize}

These findings carry broad implications for the field. If convergence is directional, then the attractor is not a generic shared structure but a specific one---language---which suggests that the discrete, compositional nature of language acts as a natural compression target. This perspective sharpens a question left open by PRH: if representations converge, \emph{where} do they converge to? Our analysis reveals that convergence is not mutual in any meaningful sense---it is asymmetric, and the asymmetry follows the abstraction hierarchy. Moreover, the phenomenon is remarkably robust: it holds across ten model families spanning four orders of magnitude in parameters and is confirmed by $k$-sensitivity analysis across multiple neighborhood sizes.

Methodologically, our work highlights a systematic blind spot: the exclusive reliance on symmetric measures in representation comparison has left convergence direction completely unexamined. We do not propose a new metric; rather, we recognize that \cycleknn's asymmetry---catalogued by~\citet{huh2024platonic} but never analyzed---encodes directional structure, and we build an analysis framework to extract it. This ``directional convergence analysis'' framework is useful independently of any specific hypothesis about the attractor.

Based on these findings, we propose the \textbf{Wittgensteinian Representation Hypothesis} (\ours): \emph{the semantic structure of language constitutes the asymptotic attractor of multimodal representation convergence.}

The name invokes Wittgenstein's dictum ``The limits of my language mean the limits of my world'' (\emph{Tractatus}~5.6), reinterpreted here as a statement about representation learning: the structural constraints imposed by language define the boundary toward which all perceptual representations converge under sufficient optimization. Unlike the Platonic Representation Hypothesis, which is agnostic about the convergence endpoint, \ours makes a falsifiable directional prediction---and our experiments confirm it across all 58 models.

The remainder of this paper is organized as follows. Section~\ref{sec:prelim} defines \cycleknn and its directional interpretation. Section~\ref{sec:empirical} presents empirical findings on directional asymmetry, intra-modality consensus, and scale invariance. Section~\ref{sec:mechanism} provides mechanistic analysis. Section~\ref{sec:wrh} formally states \ours and positions it within the hypothesis landscape, and Sections~\ref{sec:related}--\ref{sec:conclusion} discuss related work and conclude.

\section{Preliminaries}
\label{sec:prelim}

\subsection{Cycle-kNN: An Asymmetric Alignment Measure}
\label{sec:cycleknn_def}

Let $\mathbf{X} \in \mathbb{R}^{N \times d_1}$ and $\mathbf{Y} \in \mathbb{R}^{N \times d_2}$ denote representations of $N$ shared stimuli. Standard alignment measures---CKA~\citep{kornblith2019similarity}, mutual kNN~\citep{huh2024platonic}, RSA~\citep{kriegeskorte2008representational}---are symmetric by construction: $s(\mathbf{X}, \mathbf{Y}) = s(\mathbf{Y}, \mathbf{X})$. They measure alignment \emph{degree} but cannot reveal its \emph{direction}.

\cycleknn~\citep{huh2024platonic} computes alignment through a two-step nearest-neighbor cycle:
\begin{align}
\text{\cycleknn}(\mathbf{X} \to \mathbf{Y}; k) = \frac{1}{N} \sum_{i=1}^{N} \mathbf{1}\left[ i \in \mathrm{kNN}_\mathbf{X}\!\big(\mathrm{kNN}_\mathbf{Y}(i)\big) \right]
\label{eq:cycleknn}
\end{align}
where $\mathrm{kNN}_\mathbf{Y}(i)$ finds the $k$ nearest neighbors of $i$ in space $\mathbf{Y}$, and $\mathrm{kNN}_\mathbf{X}(\cdot)$ returns to space $\mathbf{X}$. \textbf{Our key observation}: although \citet{huh2024platonic} note that \cycleknn is the only asymmetric metric among eight they catalog, its asymmetry has never been analyzed. We show it carries structured directional information.

The asymmetry arises because when the first-hop space $\mathbf{Y}$ has more coherent neighborhoods, semantically similar samples cluster tightly, making the return hop more likely to succeed:
\begin{align}
\text{\cycleknn}(\mathbf{X} \!\to\! \mathbf{Y}) > \text{\cycleknn}(\mathbf{Y} \!\to\! \mathbf{X}) \;\Longrightarrow\; \mathbf{Y}\text{'s neighborhoods are more coherent.}
\end{align}
We define the \textbf{directional gap} $\Delta(\mathbf{X}, \mathbf{Y}) = \text{\cycleknn}(\mathbf{X} \to \mathbf{Y}) - \text{\cycleknn}(\mathbf{Y} \to \mathbf{X})$. A consistently positive $\Delta$ indicates that $\mathbf{Y}$'s structure is a more stable reference frame. Robustness is confirmed through synthetic experiments with matched dimensionality (Appendix~\ref{app:synthetic}), $k$-sensitivity analysis ($k = 1 \ldots 50$), and permutation tests.

\subsection{Experimental Setup}
\label{sec:setup}

\paragraph{Models.} We assemble 58 independently trained models spanning three modalities along an \emph{abstraction hierarchy}: 7 \textbf{point cloud} models (PointMAE~\citep{pang2022masked}, PointBERT~\citep{yu2022pointbert}, PointGPT~\citep{zhang2022pointclip}, ReCon++~\citep{qi2025shapellm}; 16M--300M params), 22 \textbf{vision} models across supervised, self-supervised, and contrastive paradigms (ViT~\citep{dosovitskiy2021image}, MAE~\citep{he2022masked}, DINOv2~\citep{oquab2024dinov2}, DINOv3~\citep{caron2025dinov3}, CLIP~\citep{radford2021learning}; 5.7M--7B), and 29 \textbf{language} models from five families (BLOOMZ~\citep{muennighoff2023crosslingual}, LLaMA~\citep{touvron2023llama}, Qwen2.5~\citep{qwen2025qwen25}, Qwen3~\citep{qwen2025qwen3}, InternLM~\citep{cai2024internlm2}; 560M--72B). Except for CLIP and ReCon++ (cross-modal controls), \emph{no model has seen data from another modality during training}. Full details are in Table~\ref{tab:model_full} (Appendix~\ref{app:details}).

\paragraph{Abstraction hierarchy.} The three modalities are arranged along an abstraction hierarchy: \textbf{3D point clouds} encode raw physical geometry; \textbf{2D vision} projects 3D reality onto a 2D plane, introducing viewpoint-dependent abstraction; \textbf{language} compresses perceptual experience into discrete symbolic tokens, achieving maximal abstraction.

\paragraph{Data and features.} For vision--language alignment, we use 1,024 image--text pairs from the WiT dataset~\citep{srinivasan2021wit}. For point cloud--language and point cloud--vision alignment, we use ShapeNet55~\citep{chang2015shapenet} with matched 3D objects, rendered images, and text descriptions. We extract representations from each model at every layer, yielding per-layer feature matrices $\mathbf{X}_l \in \mathbb{R}^{N \times d}$. Features are L2-normalized before computing cosine similarity for nearest-neighbor search.

\paragraph{Metric computation.} For each cross-modality model pair $(M_x, M_y)$, we compute \cycleknn in both directions at every layer pair $(l_x, l_y)$, producing a score matrix of size $L_x \times L_y$. We report the maximum score across layer pairs (best-layer alignment) as the summary statistic. We simultaneously compute symmetric measures (CKA, mutual kNN) on the same model pairs as controls. Default $k\!=\!10$.

\section{Directional Convergence: Empirical Analysis}
\label{sec:empirical}

This section presents our core empirical findings. We demonstrate that representation convergence is not symmetric---it has a consistent direction pointing toward language.

\subsection{Directional Asymmetry: Cycle-kNN Reveals Convergence Direction}
\label{sec:direction}

\begin{figure}[t]
  \centering
  \includegraphics[width=0.75\linewidth]{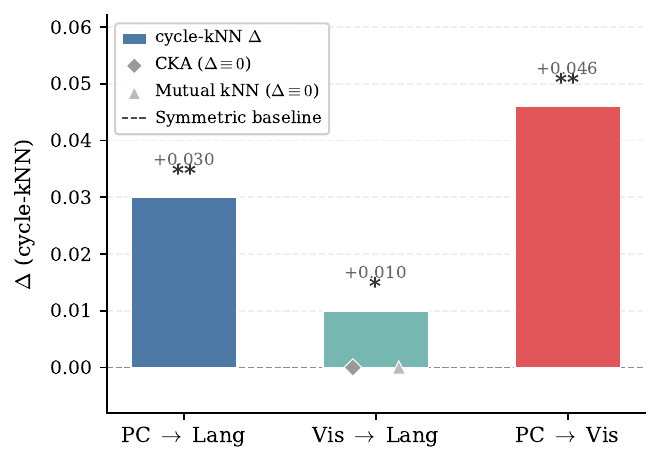}
  \caption{\textbf{Directional asymmetry across modality pairs.} Bars show the directional gap $\Delta = \text{\cycleknn}(A \to B) - \text{\cycleknn}(B \to A)$ for each cross-modality combination ($k\!=\!10$). All $\Delta > 0$, confirming convergence toward the more abstract modality. Gray markers indicate symmetric measures (CKA, mutual kNN), which yield $\Delta \equiv 0$ by construction on the same model pairs. ${}^{**}p < 0.01$, ${}^{*}p < 0.05$ (permutation test, $n\!=\!1000$).}
  \label{fig:asymmetry}
\end{figure}

We compute \cycleknn in both directions for all cross-modality model pairs across the six directional combinations: language $\to$ vision, vision $\to$ language, language $\to$ point cloud, point cloud $\to$ language, vision $\to$ point cloud, and point cloud $\to$ vision (Figure~\ref{fig:asymmetry}).

\textbf{Finding 1.} \emph{Across all model pairs, convergence exhibits a consistent directional asymmetry: non-language modalities align toward language more than language aligns toward them.}

\begin{table}[t]
  \caption{\textbf{Directional asymmetry across modality pairs.} Mean $\pm$ std of \cycleknn in each direction ($k\!=\!10$), computed over all model pairs. The directional gap $\Delta > 0$ consistently indicates convergence toward the more abstract modality. Permutation test ($n\!=\!1000$) confirms significance. Symmetric measures (CKA, mutual kNN) show $\Delta \approx 0$ on the same pairs.}
  \label{tab:direction}
  \centering
  \small
  \setlength{\tabcolsep}{4pt}
  \begin{tabular}{@{}l cc c c@{}}
  \toprule
  \textbf{Direction Pair} & \makecell{\cycleknn\\($A\!\to\!B$)} & \makecell{\cycleknn\\($B\!\to\!A$)} & $\Delta$ & $p$-value \\
  \midrule
  Point Cloud $\to$ Language   & 0.628 $\pm$ 0.091 & 0.598 $\pm$ 0.098 & +0.030 & $<$0.01 \\
  Vision $\to$ Language        & 0.577 $\pm$ 0.075 & 0.566 $\pm$ 0.072 & +0.010 & $<$0.05 \\
  Point Cloud $\to$ Vision     & 0.720 $\pm$ 0.134 & 0.674 $\pm$ 0.133 & +0.046 & $<$0.01 \\
  \midrule
  \multicolumn{5}{@{}l}{\emph{Symmetric controls (same model pairs):}} \\
  \midrule
  CKA: Vis-Lang                  & \multicolumn{2}{c}{0.410 $\pm$ 0.066} & $\approx 0$ & -- \\
  Mutual kNN: Vision--Language & \multicolumn{2}{c}{0.385 $\pm$ 0.071} & $\approx 0$ & -- \\
  \bottomrule
  \end{tabular}
\end{table}

\begin{figure*}[t]
  \centering
  \includegraphics[width=\linewidth]{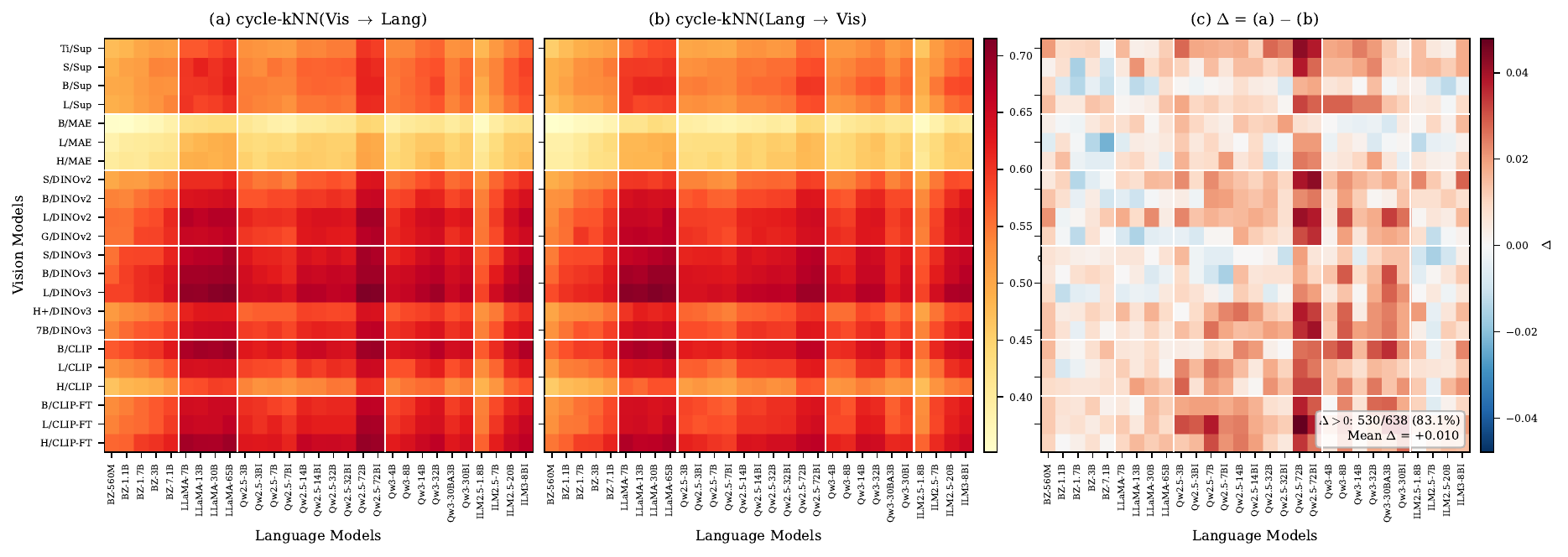}
  \caption{\textbf{Systematic directional asymmetry across all model pairs.}
    (a)~\cycleknn(Vision $\to$ Language) and (b)~\cycleknn(Language $\to$ Vision) score matrices for all 22 vision $\times$ 29 language model pairs ($k\!=\!10$, WiT-1024 dataset). Both panels share the same color scale. Panel~(a) is systematically brighter than~(b), confirming that Language neighborhoods are more coherent when probed from Vision. (c)~Element-wise difference $\Delta_{ij}$: 530/638 pairs (83.1\%) have $\Delta > 0$ (mean $\Delta = +0.010$), ruling out the asymmetry being driven by outlier models.}
  \label{fig:heatmap}
\end{figure*}

\paragraph{Robustness.}
We verify that the directional asymmetry is robust across $k = 1, 2, 3, 5, 10, 20, 50$---the sign of $\Delta$ is consistent across all $k$ values (Appendix~\ref{app:results}). Permutation tests ($n\!=\!1000$) confirm statistical significance ($p < 0.01$ for all modality pairs; Table~\ref{tab:direction}). On the same model pairs, CKA and mutual kNN yield $\Delta \approx 0$, confirming that directionality is invisible to symmetric analysis. The full pairwise score matrices (Figure~\ref{fig:heatmap}) confirm this is systematic across all 638 model pairs.

\subsection{Intra-Modality Consensus: Language Representations Agree Most}
\label{sec:consensus}

\textbf{Finding 2.} \emph{Language models exhibit the highest intra-modality representational consensus (mean CKA = 0.904), followed by vision (0.857), with point cloud showing the lowest (0.440).}

\begin{figure*}[t]
  \centering
  \includegraphics[width=\linewidth]{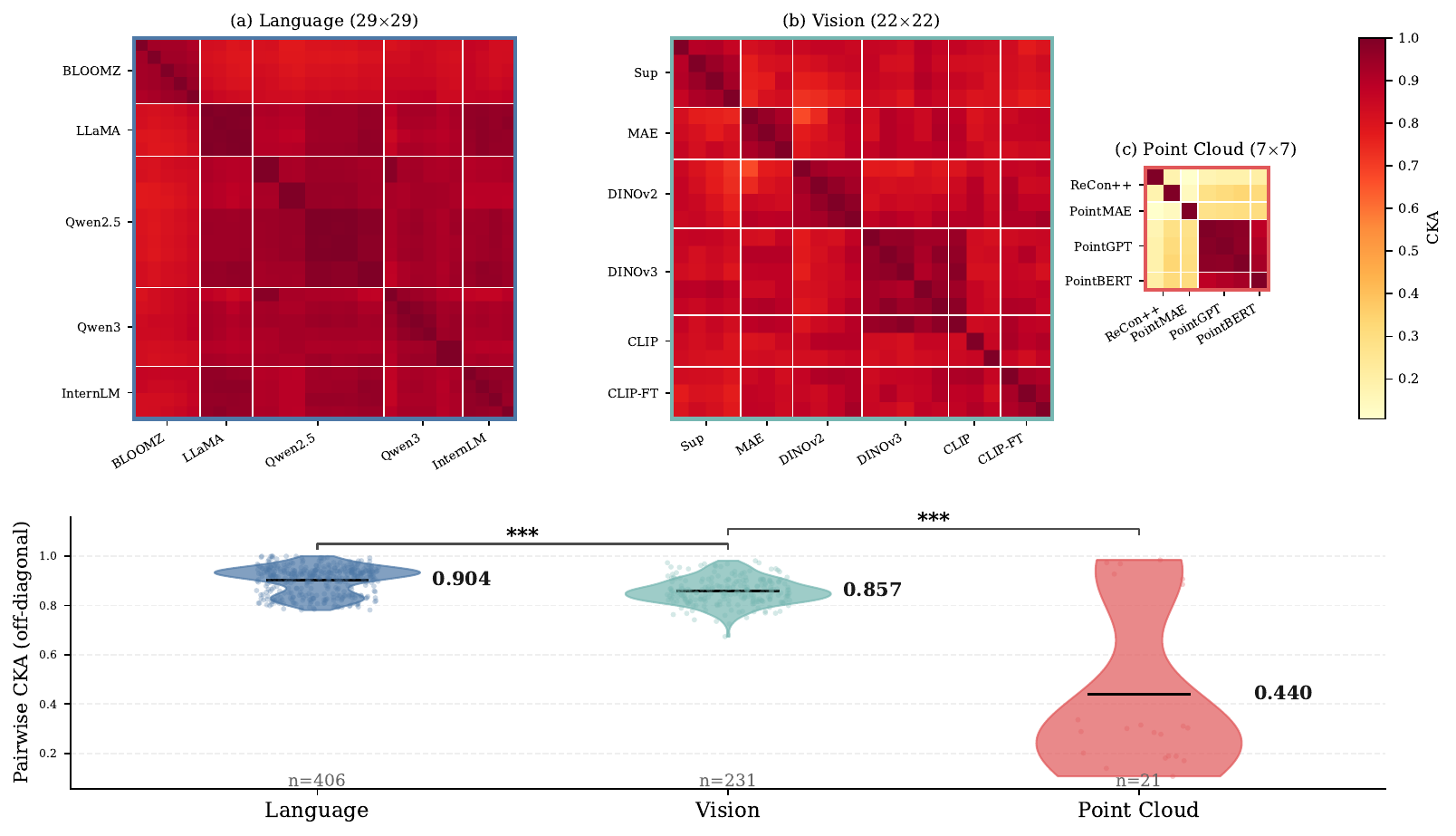}
  \caption{\textbf{Intra-modality representational consensus.}
    Top: pairwise CKA heatmaps (shared color scale).
    Bottom: violin plots confirm ordering Language $>$ Vision $>$ Point Cloud ($p < 0.001$, Mann--Whitney $U$). Consistent with language being the convergence attractor.}
  \label{fig:consensus}
\end{figure*}

We compute pairwise alignment within each modality using both CKA and mutual kNN. Figure~\ref{fig:consensus} and Table~\ref{tab:consensus} show the results: language models exhibit uniformly high agreement, while point cloud models show high variance---consistent with the attractor interpretation that representations near the attractor exhibit low variance.

\begin{table}[t]
  \caption{\textbf{Intra-modality consensus.} Mean $\pm$ std of pairwise alignment within each modality.}
  \label{tab:consensus}
  \centering
  \small
  \begin{tabular}{@{}l c c c@{}}
  \toprule
  \textbf{Modality} & \textbf{\#Pairs} & \textbf{CKA} & \textbf{Mutual kNN} \\
  \midrule
  Language     & 406 & $0.904 \pm 0.055$ & $0.562 \pm 0.100$ \\
  Vision       & 231 & $0.857 \pm 0.051$ & $0.502 \pm 0.079$ \\
  Point Cloud  &  21 & $0.440 \pm 0.324$ & $0.248 \pm 0.236$ \\
  \bottomrule
  \end{tabular}
\end{table}

\subsection{Directionality Is Robust Across Scales}
\label{sec:scaling}

\textbf{Finding 3.} \emph{The directional asymmetry $\Delta > 0$ toward language holds for every vision model (22/22), every point cloud model (7/7), and nearly every language model (28/29 for Vision--Language, 29/29 for PC--Language), regardless of scale, architecture, or training paradigm.}

\begin{figure}[t]
  \centering
  \includegraphics[width=\linewidth]{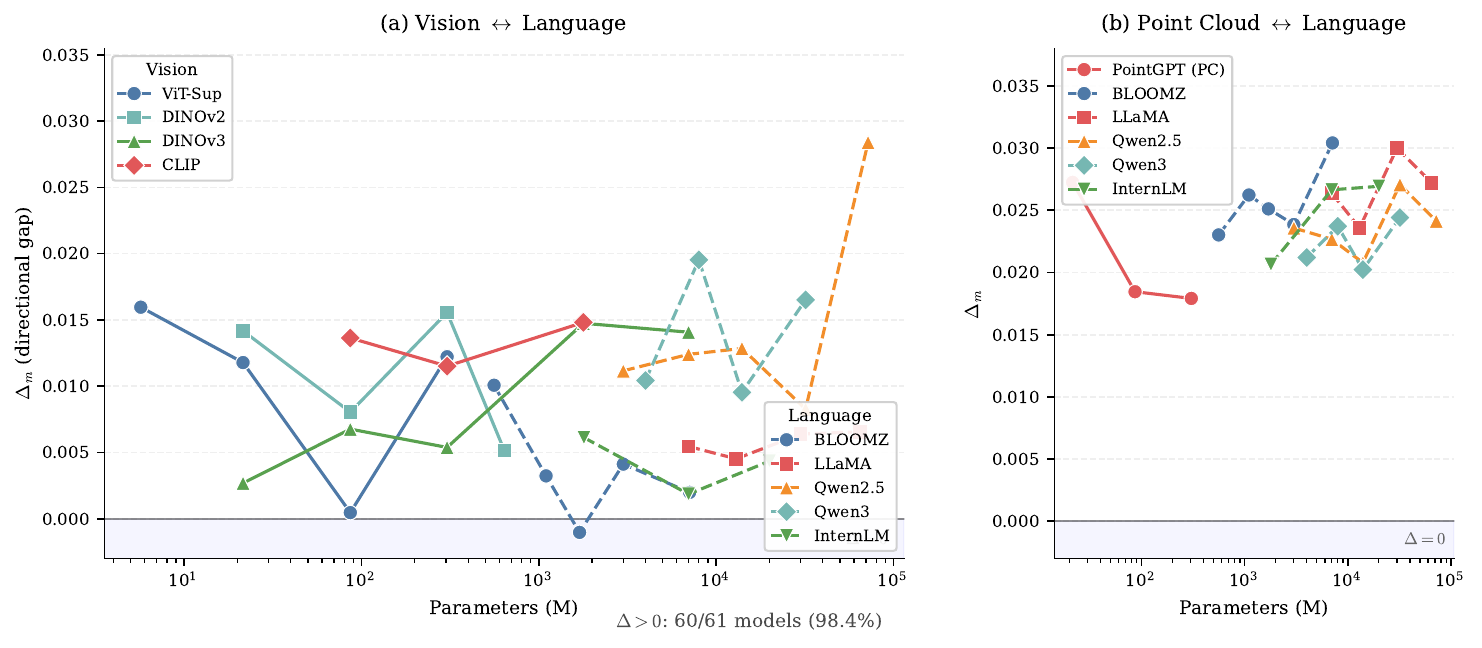}
  \caption{\textbf{Scale-invariant directionality.}
    Per-model $\Delta_m$ vs.\ parameter count for ten model families.
    (a)~Vision$\leftrightarrow$Language. (b)~PC$\leftrightarrow$Language.
    60/61 combinations (98.4\%) have $\Delta > 0$, confirming scale-invariance.}
  \label{fig:scaling}
\end{figure}

For each model $m$ of modality $A$, we compute $\Delta_m = \overline{\mathrm{cyc\text{-}kNN}}(A_m \to B) - \overline{\mathrm{cyc\text{-}kNN}}(B \to A_m)$ averaged over all target-modality models. Figure~\ref{fig:scaling} plots $\Delta_m$ vs.\ parameter count for ten families spanning four orders of magnitude (5.7M--72B). While neural scaling laws~\citep{kaplan2020scaling} predict smooth capability gains with scale, the directional asymmetry shows no such dependence---$\Delta$ remains positive regardless of parameter count.

\section{Why Language? Mechanistic Analysis}
\label{sec:mechanism}

Having established the empirical phenomenon, we investigate its underlying mechanism. We propose that the directionality is associated with \textbf{feature density asymmetry} and that this asymmetry has a principled interpretation under the Information Bottleneck framework.

\subsection{Feature Density Differs Across Modalities}
\label{sec:density}

We measure representational compactness using pairwise mean distance:
\begin{align}
D(\mathbf{X}) = \frac{2}{N(N-1)} \sum_{i < j} \| \mathbf{x}_i - \mathbf{x}_j \|
\end{align}
Lower $D$ indicates more compact representations (samples cluster more tightly).

\textbf{Finding 4.} \emph{At alignment-optimal layers, language representations are among the most compact. Layer-wise density profiles reveal characteristic modality-specific patterns that correlate with the observed convergence directionality.}

\begin{figure}[t]
  \centering
  \includegraphics[width=\linewidth]{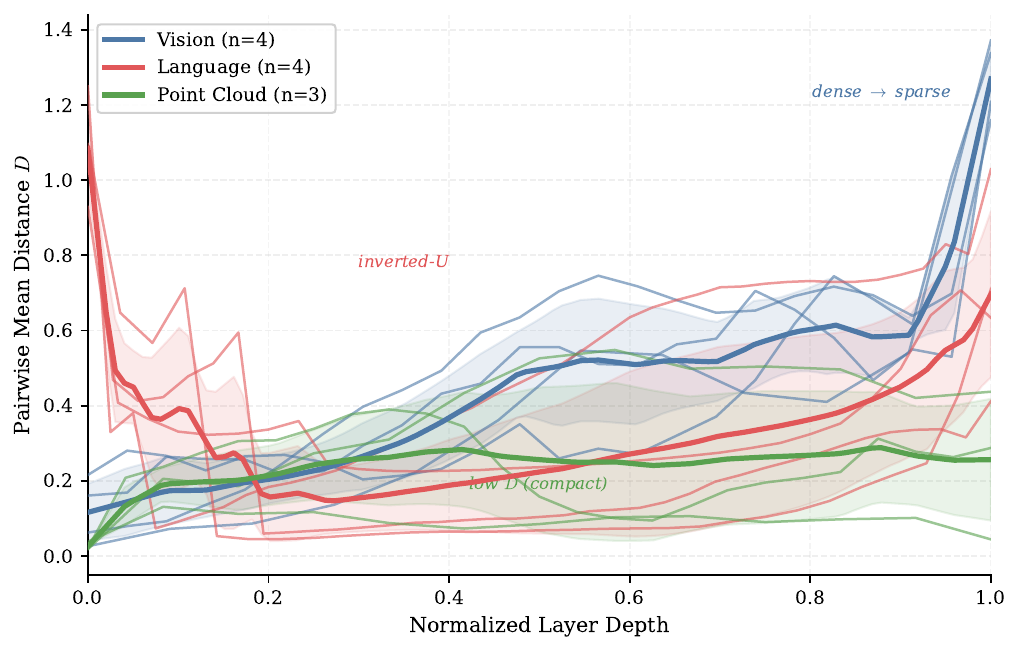}
  \caption{\textbf{Layer-wise representational density.}
    Pairwise mean distance $D$ (on $\ell_2$-normalised features) across normalised layer depth for representative models from each modality.
    Vision models (blue) show monotonically increasing $D$ (dense$\to$sparse);
    language models (red) follow an inverted-U pattern, reaching maximum
    compactness at later layers;
    point cloud models (green) show variable density patterns.
    Bold lines: per-modality mean; shaded regions: $\pm$1 s.d.}
  \label{fig:density_layers}
\end{figure}

Specifically, vision models show $D$ increasing monotonically across layers (early layers are dense, later layers sparse); language models follow an inverted-U trajectory where $D$ first increases then decreases, reaching minimal density at intermediate-to-late layers---precisely the layers where cross-modal alignment peaks (Figure~\ref{fig:density_layers}). These modality-specific patterns are consistent across model families, suggesting they reflect intrinsic properties of the input modality rather than architecture-specific artifacts.

The key observation is that language models achieve maximal compactness at the same layers where cross-modal alignment is strongest, suggesting a geometric link: the coherent neighborhood structure of compact representations facilitates cycle-kNN return, producing the observed directional asymmetry. Controlled synthetic experiments (Appendix~\ref{app:synthetic}) confirm that density asymmetry alone is sufficient to produce \cycleknn directionality in matched-dimensionality settings.

\subsection{Information Bottleneck Interpretation}
\label{sec:ib}

The Information Bottleneck (IB) framework~\citep{tishby2000information, tishby2015deep} provides a principled interpretive lens for why language representations tend to be compact. The IB objective for a representation $\mathbf{Z}$ of input $\mathbf{X}$ for task $\mathbf{Y}$ is:
\begin{align}
\min_{p(\mathbf{z}|\mathbf{x})} I(\mathbf{X}; \mathbf{Z}) - \beta \, I(\mathbf{Z}; \mathbf{Y})
\end{align}
where $I(\cdot;\cdot)$ denotes mutual information. The first term encourages compression; the second preserves task-relevant information. As compression strengthens, optimal representations undergo phase transitions toward increasingly discrete, combinatorial structures~\citep{strouse2017deterministic}---precisely the structural form of natural language.

\paragraph{Interpretive account.} Natural language can be viewed as humanity's most optimized compression of perceptual experience. Under the IB lens, a sufficiently trained model---regardless of input modality---is driven toward maximal compression of task-irrelevant information, converging on a structure that mirrors language. Higher compression implies lower representational entropy, manifesting geometrically as tighter clustering (lower $D$), which in turn predicts a positive directional gap $\Delta > 0$ when language serves as the first-hop reference space. This is consistent with our empirical observations.

This interpretation connects to multiple independent theoretical lines: neural thermodynamics~\citep{ziyin2025neural}, which characterizes training dynamics as progressive symmetry-breaking driven by entropic forces; evidence that natural datasets exhibit universal statistical structure~\citep{levi2025rmt} whose power-law scaling may independently drive convergence; and \citet{schrodi2025modality}, who demonstrate that information imbalance between modalities drives both the modality gap and representation structure. Recent work directly applying IB regularization to multimodal alignment~\citep{almudevar2025ib} confirms that explicitly constraining modality-specific information improves cross-modal convergence, while \citet{gui2025multimodal} show that contrastive learning adapts representations to the intrinsic dimension of shared latent variables. The information asymmetry at the data level manifests as density asymmetry at the representation level, consistent with the directional convergence we observe.

\section{The Wittgensteinian Representation Hypothesis}
\label{sec:wrh}

Drawing together the empirical findings (Section~\ref{sec:empirical}) and mechanistic analysis (Section~\ref{sec:mechanism}), we formally state:

\begin{quote}
\textbf{The Wittgensteinian Representation Hypothesis (\ours).} \emph{The semantic structure of language constitutes the asymptotic attractor of multimodal representation convergence. As independently trained models become sufficiently powerful, their learned representations---regardless of input modality---converge toward the representational geometry defined by language's semantic structure.}
\end{quote}

The name invokes Wittgenstein's observation in the \emph{Tractatus}~\citep{wittgenstein1921tractatus}: ``Die Grenzen meiner Sprache bedeuten die Grenzen meiner Welt.'' In representation learning terms, language provides a \emph{structural schema} toward which cross-modal representations converge. Table~\ref{tab:comparison} positions \ours within the hypothesis landscape.

\begin{table}[H]
  \caption{Comparison of representation convergence hypotheses.}
  \label{tab:comparison}
  \centering
  \footnotesize
  \setlength{\tabcolsep}{4pt}
  \begin{tabular}{@{}lllll@{}}
    \toprule
    \textbf{Hypothesis} & \textbf{Endpoint} & \textbf{Direction} & \textbf{Scope} & \textbf{Method} \\
    \midrule
    PRH \citep{huh2024platonic}             & Stat.\ of reality    & --            & Inter-model & Symmetric \\
    Sem.\ Hub \citep{wu2025semantic}        & Lang.\ internal hub  & Intra-model   & Intra-model & Causal int. \\
    Indra \citep{lu2025indra}               & Relational struct.   & --            & Inter-model & Symmetric \\
    UNE \citep{tasker2026une}               & Gaussian geometry    & --            & Inter-model & Symmetric \\
    Neural Thermo.\ \citep{ziyin2025neural} & Symmetry-constr.     & --            & Inter-model & Theoretical \\
    \textbf{\ours (Ours)}                   & \textbf{Lang.\ struct.} & \textbf{Toward lang.} & \textbf{Inter-model} & \textbf{Asymmetric} \\
    \bottomrule
  \end{tabular}
\end{table}

\ours differs from PRH in providing an \emph{operationally testable} endpoint (language representations) rather than an unobservable ``statistical structure of reality''; from Semantic Hub~\citep{wu2025semantic} in addressing inter-model (not intra-model) convergence; and from Indra/UNE in employing asymmetric rather than symmetric tools.

\paragraph{Empirical support.}
Our experiments provide four converging lines of evidence for \ours: (1)~directional asymmetry toward language across all three modality pairs (Section~\ref{sec:direction}); (2)~highest intra-modality consensus for language, consistent with attractor proximity (Section~\ref{sec:consensus}); (3)~scale-invariance of the directionality across 58 models spanning four orders of magnitude (Section~\ref{sec:scaling}); and (4)~a mechanistic link via feature density asymmetry, where language's compactness explains why \cycleknn preferentially succeeds when language serves as the first-hop space (Section~\ref{sec:mechanism}). A supplementary exploration with directional CKA shows directionally consistent results on the vision--language axis, though this metric has known limitations when feature dimensions differ across modalities (Appendix~\ref{app:dcka}).

\section{Related Work}
\label{sec:related}

\paragraph{Representation convergence.}
The PRH~\citep{huh2024platonic} posits convergence toward a shared statistical model of reality, with subsequent formalizations including Indra representations~\citep{lu2025indra}, Universal Normal Embeddings~\citep{tasker2026une}, Neural Thermodynamics~\citep{ziyin2025neural}, and learning-theoretic analyses~\citep{insulla2025learning, luthra2026alignment}. Empirically, frozen unimodal encoders can be bridged by lightweight projections~\citep{maniparambil2025frozen, groger2025structure}, blind matching without paired data achieves non-trivial accuracy~\citep{schnaus2025blind, yacobi2025shared}, and generative pretraining produces implicit cross-modal alignment~\citep{xiao2025scaling}. A key limitation: all rely exclusively on symmetric measures, leaving convergence direction unexamined.

\paragraph{Asymmetric and directional analysis.}
\citet{acevedo2025connecting} applied Information Imbalance~\citep{glielmo2022ranking} to independently trained models, observing text$\to$vision asymmetry, but without proposing a directional hypothesis or mechanistic explanation. The Semantic Hub Hypothesis~\citep{wu2025semantic} describes language as a routing hub \emph{within} multimodal models---complementary to our inter-model analysis. Recent surveys~\citep{koutoupis2026confu, tjandrasuwita2025emergence} note the open question of whether a specific modality should serve as the alignment anchor; our work provides the first systematic evidence.

\paragraph{Boundary conditions.}
Convergence is not unconditional: alignment depends on shared information structure~\citep{tjandrasuwita2025emergence}, training objective is the primary determinant of representational consistency~\citep{ciernik2025objective}, and early-training perturbations can permanently alter representations~\citep{altintas2025butterfly}. At the measurement level, distributional closeness does not guarantee representational similarity~\citep{nielsen2025closeness, smith2025functional}, and \citet{kabra2026omnivorous} find near-random cross-modal similarity for certain modality pairs. These boundaries underscore the importance of our direct geometric measurement via \cycleknn.

\section{Discussion and Conclusion}
\label{sec:conclusion}

\paragraph{Implications of directional convergence.}
The Platonic Representation Hypothesis identified a profound phenomenon: independently trained models are converging. But the question of \emph{where} they converge has remained unanswered, obscured by the exclusive use of symmetric measures. Our directional convergence analysis reveals that convergence is not symmetric---it has a consistent direction toward language. This suggests that the attractor is not a generic ``statistical structure of reality'' but something more specific: the compressed, compositional structure that language imposes on experience.

\paragraph{Methodological contribution.}
Beyond the specific hypothesis, our work highlights a systematic blind spot: the exclusive reliance on symmetric measures has left convergence direction entirely unexamined. We do not propose a new metric; rather, we recognize that \cycleknn's directional asymmetry encodes structured information and build a framework to extract it.

\paragraph{Limitations and future work.}
Our abstraction hierarchy covers three modalities with 58 models; extending to audio, tactile, or code modalities would strengthen generality. The IB-based mechanistic account is an interpretive framework rather than a formal proof. Whether fundamentally different architectures could bypass the language attractor remains open, though our exclusive use of unimodal models mitigates language data leakage concerns. Future directions include multi-metric validation with additional asymmetric measures, tracking directionality during training dynamics, and formal theoretical connections between IB optimality and linguistic structure.

\bibliographystyle{plainnat}
\bibliography{references}

\clearpage
\appendix
\raggedbottom

{\Large\textbf{Appendix}}

\section{Cycle-kNN Is Inherently Asymmetric}
\label{app:asymmetry_proof}

We prove that \cycleknn is asymmetric for $k \geq 2$, the regime used throughout this paper ($k{=}10$).

\paragraph{Definition.}
Let $\mathbf{X}, \mathbf{Y} \in \mathbb{R}^{N \times d}$ be two representation matrices of $N$ shared stimuli.
For $k \geq 1$:
\begin{align}
\text{\cycleknn}(\mathbf{X} \to \mathbf{Y}; k) = \frac{1}{N} \sum_{i=1}^{N} \mathbf{1}\!\left[ i \in \bigcup_{j \in \mathrm{kNN}_\mathbf{Y}(i)} \mathrm{kNN}_\mathbf{X}(j) \right]
\end{align}
where $\mathrm{kNN}_\mathbf{Y}(i)$ denotes the $k$ nearest neighbors of point~$i$ in space~$\mathbf{Y}$ (excluding~$i$ itself), and similarly for $\mathrm{kNN}_\mathbf{X}$.

\paragraph{Proposition (symmetry at $k{=}1$).}
For $k{=}1$ with unique nearest neighbors, $\text{\cycleknn}(\mathbf{X}{\to}\mathbf{Y}; 1) = \text{\cycleknn}(\mathbf{Y}{\to}\mathbf{X}; 1)$ always holds.

\emph{Proof.}
Let $A = \{i : \mathrm{NN}_\mathbf{X}(\mathrm{NN}_\mathbf{Y}(i)) = i\}$ and $B = \{i : \mathrm{NN}_\mathbf{Y}(\mathrm{NN}_\mathbf{X}(i)) = i\}$.
Define $\varphi: A \to B$ by $\varphi(i) = \mathrm{NN}_\mathbf{Y}(i)$.
If $i \in A$, let $j = \mathrm{NN}_\mathbf{Y}(i)$; then $\mathrm{NN}_\mathbf{X}(j) = i$, so $\mathrm{NN}_\mathbf{Y}(\mathrm{NN}_\mathbf{X}(j)) = \mathrm{NN}_\mathbf{Y}(i) = j$, hence $j \in B$.
Injectivity: if $\varphi(i_1) = \varphi(i_2) = j$, then $\mathrm{NN}_\mathbf{X}(j) = i_1 = i_2$.
Surjectivity: for $j \in B$, set $i = \mathrm{NN}_\mathbf{X}(j)$; then $\mathrm{NN}_\mathbf{Y}(i) = j$ and $\mathrm{NN}_\mathbf{X}(j) = i$, so $i \in A$ with $\varphi(i) = j$.
Thus $\varphi$ is a bijection, $|A| = |B|$, and the two scores coincide. \hfill$\square$

\paragraph{Theorem (asymmetry at $k \geq 2$).}
For $k \geq 2$, there exist configurations where $\text{\cycleknn}(\mathbf{X}{\to}\mathbf{Y}; k) \neq \text{\cycleknn}(\mathbf{Y}{\to}\mathbf{X}; k)$.

\emph{Proof by construction.}
Let $N{=}6$, $d{=}1$, $k{=}2$ with
$\mathbf{X} = (15, 26, 49, 60, 87, 90)^\top$ and
$\mathbf{Y} = (34, 56, 58, 57, 63, 37)^\top$.
All pairwise distances in both spaces are distinct at the $k$/$k{+}1$ boundary, eliminating tie-breaking ambiguity.

\emph{Direction $\mathbf{X} \to \mathbf{Y}$ ($k{=}2$):}
For each $i$, we compute $S_i = \mathrm{2NN}_\mathbf{Y}(i)$ and the return set $R_i = \bigcup_{j \in S_i} \mathrm{2NN}_\mathbf{X}(j)$:
\begin{align*}
i{=}1:\; & S_1 = \{6,2\},\; R_1 = \mathrm{2NN}_\mathbf{X}(6) \cup \mathrm{2NN}_\mathbf{X}(2) = \{5,4\} \cup \{1,3\} = \{1,3,4,5\} \ni 1 \;\checkmark \\
i{=}2:\; & S_2 = \{4,3\},\; R_2 = \{3,5\} \cup \{4,2\} = \{2,3,4,5\} \ni 2 \;\checkmark \\
i{=}3:\; & S_3 = \{4,2\},\; R_3 = \{3,5\} \cup \{1,3\} = \{1,3,5\} \ni 3 \;\checkmark \\
i{=}4:\; & S_4 = \{2,3\},\; R_4 = \{1,3\} \cup \{4,2\} = \{1,2,3,4\} \ni 4 \;\checkmark \\
i{=}5:\; & S_5 = \{3,4\},\; R_5 = \{4,2\} \cup \{3,5\} = \{2,3,4,5\} \ni 5 \;\checkmark \\
i{=}6:\; & S_6 = \{1,2\},\; R_6 = \{2,3\} \cup \{1,3\} = \{1,2,3\} \not\ni 6 \;\times
\end{align*}
Score: $\text{\cycleknn}(\mathbf{X}{\to}\mathbf{Y}; 2) = 5/6$.

\emph{Direction $\mathbf{Y} \to \mathbf{X}$ ($k{=}2$):}
\begin{align*}
i{=}1:\; & S_1 = \{2,3\},\; R_1 = \mathrm{2NN}_\mathbf{Y}(2) \cup \mathrm{2NN}_\mathbf{Y}(3) = \{4,3\} \cup \{4,2\} = \{2,3,4\} \not\ni 1 \;\times \\
i{=}2:\; & S_2 = \{1,3\},\; R_2 = \{6,2\} \cup \{4,2\} = \{2,4,6\} \ni 2 \;\checkmark \\
i{=}3:\; & S_3 = \{4,2\},\; R_3 = \{2,3\} \cup \{4,3\} = \{2,3,4\} \ni 3 \;\checkmark \\
i{=}4:\; & S_4 = \{3,5\},\; R_4 = \{4,2\} \cup \{3,4\} = \{2,3,4\} \ni 4 \;\checkmark \\
i{=}5:\; & S_5 = \{6,4\},\; R_5 = \{1,2\} \cup \{2,3\} = \{1,2,3\} \not\ni 5 \;\times \\
i{=}6:\; & S_6 = \{5,4\},\; R_6 = \{3,4\} \cup \{2,3\} = \{2,3,4\} \not\ni 6 \;\times
\end{align*}
Score: $\text{\cycleknn}(\mathbf{Y}{\to}\mathbf{X}; 2) = 3/6 = 1/2$.

Since $5/6 \neq 1/2$, the metric is asymmetric. \hfill$\square$

\paragraph{Why $k \geq 2$ breaks symmetry.}
The bijection proof for $k{=}1$ relies on the fact that $\mathrm{NN}$ returns a \emph{single} point, creating a one-to-one pairing between success sets.
For $k \geq 2$, the union $\bigcup_{j \in S_i} \mathrm{kNN}_\mathbf{X}(j)$ aggregates return paths from multiple intermediate points.
A space with compact clusters (small pairwise distances, like $\mathbf{X}$ above where $\{87,90\}$ and $\{15,26\}$ form tight pairs) concentrates its $k$-neighborhoods on nearby points, creating redundant return paths that benefit queries \emph{originating} in the other space.
Conversely, a space with dispersed structure (like $\mathbf{Y}$ above where points 2--5 span only the range $[56,63]$) offers fewer distinct return paths.
This structural asymmetry---which space provides more coherent neighborhoods for the return hop---is the geometric basis of the directional convergence signal exploited throughout this paper (using $k{=}10$, as confirmed empirically in Appendix~\ref{app:synthetic}).

\section{Relationship Between Feature Density and Cycle-kNN Directionality}
\label{app:density_theory}

We formalize the intuition connecting feature density to \cycleknn directionality.

\paragraph{Definition (Pairwise mean distance).}
For a representation matrix $\mathbf{X} \in \mathbb{R}^{N \times d}$ with L2-normalized rows:
\begin{align}
D(\mathbf{X}) = \frac{2}{N(N-1)} \sum_{i < j} \| \mathbf{x}_i - \mathbf{x}_j \|_2
\end{align}

\paragraph{Density--directionality connection.}
Lower $D(\mathbf{X})$ means samples are more tightly clustered (compact representation).
In a compact space, $k$-nearest neighbors of a query point are more likely to be \emph{semantically} similar (sharing the same ground-truth category), because the ratio of intra-class to inter-class distance is smaller.

Let $p_\text{sem}(\mathbf{X}; k)$ denote the probability that a random $k$-NN query in space $\mathbf{X}$ returns a semantically consistent neighbor.
Then:
\begin{align}
D(\mathbf{X}) < D(\mathbf{Y}) \;\Longrightarrow\; p_\text{sem}(\mathbf{X}; k) > p_\text{sem}(\mathbf{Y}; k)
\end{align}

Since \cycleknn success requires both hops to preserve semantic identity:
\begin{align}
\text{\cycleknn}(\mathbf{Y} \to \mathbf{X}) &\propto p_\text{sem}(\mathbf{X}; k) \cdot p_\text{ret}(\mathbf{Y}; k) \\
\text{\cycleknn}(\mathbf{X} \to \mathbf{Y}) &\propto p_\text{sem}(\mathbf{Y}; k) \cdot p_\text{ret}(\mathbf{X}; k)
\end{align}
where $p_\text{ret}$ denotes the return probability in the second-hop space.
When $\mathbf{X}$ is the first-hop space (compact), semantically coherent neighbors are selected, boosting the return probability.
This yields:
\begin{align}
D(\mathbf{X}) < D(\mathbf{Y}) \;\Longrightarrow\; \text{\cycleknn}(\mathbf{Y} \to \mathbf{X}) > \text{\cycleknn}(\mathbf{X} \to \mathbf{Y}) \;\Longrightarrow\; \Delta(\mathbf{Y}, \mathbf{X}) > 0
\end{align}

\paragraph{Prediction.}
Given the empirically observed density ordering $D(\text{language}) < D(\text{vision}) < D(\text{point cloud})$, the theory predicts that $\Delta > 0$ for all cross-modality pairs where the target modality has lower $D$ (more compact):
\begin{align}
\Delta(\text{pc}, \text{lang}) > 0, \quad \Delta(\text{vis}, \text{lang}) > 0, \quad \Delta(\text{pc}, \text{vis}) > 0
\end{align}
which is confirmed in Section~\ref{sec:direction}. The magnitude of each $\Delta$ is proportional to the density difference $|D(\mathbf{X}) - D(\mathbf{Y})|$ at alignment-optimal layers of specific model pairs, as verified by the correlation analysis.

\section{Synthetic Experiment Details}
\label{app:synthetic}

\paragraph{Synthetic generators.}
We use eight generators to produce paired feature spaces $(\mathbf{X}, \mathbf{Y})$ with controlled density asymmetry:
(1)~\textbf{Gaussian clusters}---$C$ clusters with shared centers, variance ratio controls density difference;
(2)~\textbf{Concentric rings}---points on rings of varying radii, ring width controls density;
(3)~\textbf{Uniform grid}---points on a regular grid with Gaussian jitter, jitter magnitude controls density;
(4)~\textbf{Uniform disk}---points uniformly sampled from a disk, radius ratio controls density;
(5)~\textbf{Swiss roll}---points on a Swiss roll manifold, noise amplitude controls density;
(6)~\textbf{S-curve}---points on an S-curve, scaling factor controls density;
(7)~\textbf{Folded manifold}---a flat manifold with controlled folding, fold amplitude controls local density;
(8)~\textbf{High-dimensional Gaussian}---isotropic Gaussians in $d = 128$ dimensions, variance ratio controls density.

\begin{figure}[ht]
\centering
\includegraphics[width=\textwidth]{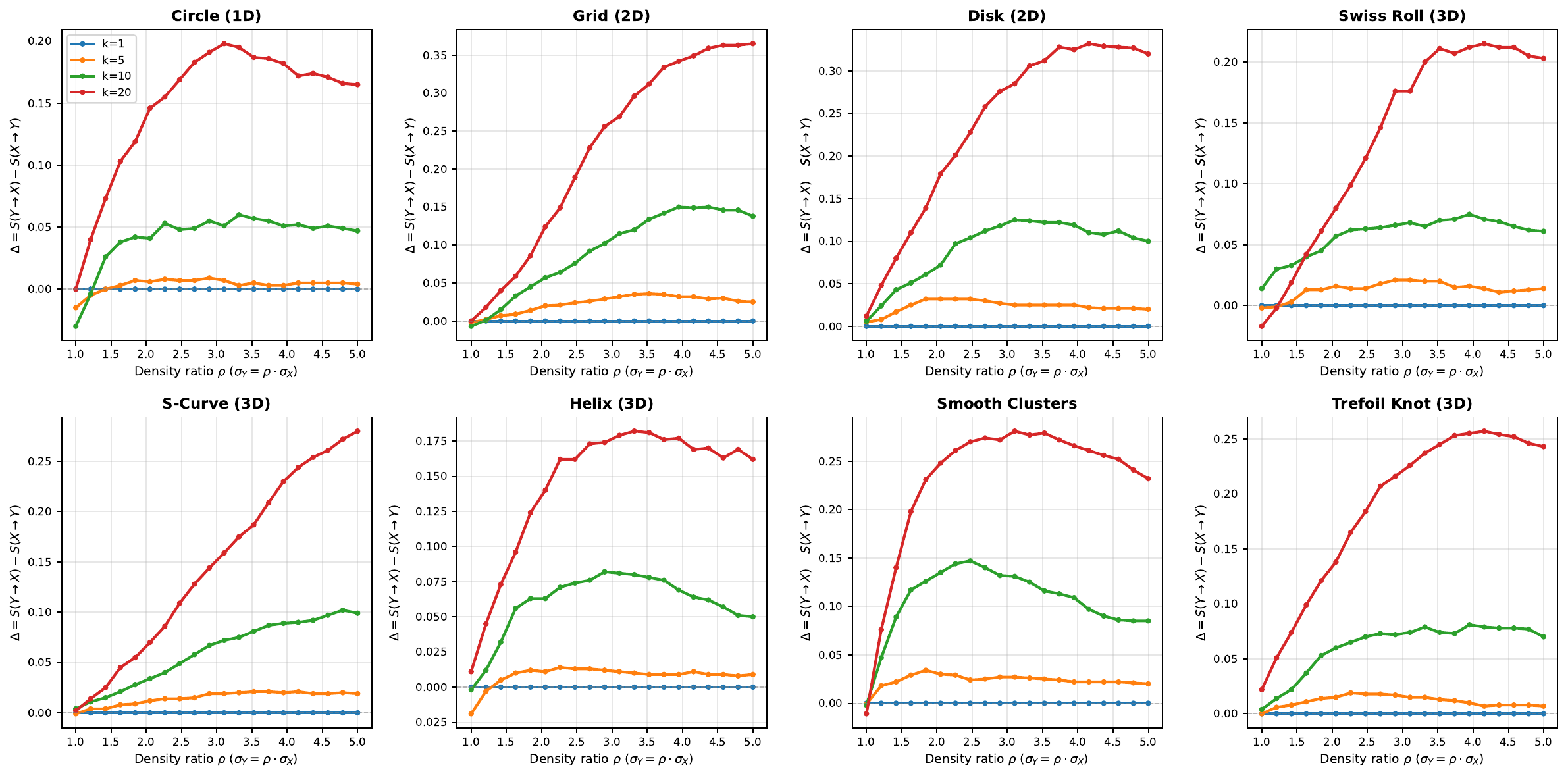}
\caption{\textbf{Synthetic validation: $\Delta$ increases monotonically with density ratio $\rho$.}
Each panel shows a different manifold generator (8 types spanning 1D--3D intrinsic dimensionality).
$X$ is a compact reference ($\sigma_{\text{base}}$ noise) and $Y$ is dispersed with noise scaled by $\rho \in [1, 5]$.
All curves confirm that $\Delta = S(Y{\to}X) - S(X{\to}Y) > 0$ once $\rho > 1$, and increases monotonically,
validating that \cycleknn correctly detects asymmetric neighborhood density.
Higher $k$ values amplify the gap; $k{=}1$ is degenerate (binary cycle match).}
\label{fig:synthetic_delta}
\end{figure}

\paragraph{Protocol.}
For each generator, we produce 20 density ratios $\rho \in [1.0, 5.0]$ (where $\rho = D(\mathbf{Y}) / D(\mathbf{X}) > 1$ means $\mathbf{X}$ is more compact).
Both spaces share $N = 1000$ samples, $d = 128$ dimensions, and semantic labels (same point indices correspond to matched samples).
We compute \cycleknn in both directions with $k \in \{1, 5, 10, 20\}$ and report $\Delta$ as a function of $\rho$.
Figure~\ref{fig:synthetic_delta} shows the results.

\section{Implementation Details}
\label{app:details}

\subsection{Feature Extraction}

\paragraph{\cycleknn computation.}
For each model pair, features are extracted at every layer. Features are L2-normalized before computing cosine similarity matrices. Nearest neighbors are found via exhaustive search (no approximate NN). $k=10$ is used by default; full $k$-sensitivity results are in Appendix~\ref{app:results}.

\paragraph{Language models.}
For autoregressive language models, we extract hidden states at the last token position (causal attention makes this the most contextually rich representation). For each layer $l$, the feature matrix is $\mathbf{X}_l \in \mathbb{R}^{N \times d_\text{hidden}}$.

\paragraph{Vision models.}
For ViT-based models, we extract the [CLS] token representation at each transformer block. For models without a [CLS] token (e.g., MAE), we use spatial average pooling.

\paragraph{Point cloud models.}
For point cloud transformers, we extract the global representation at each block (typically obtained via average pooling over point tokens).

\begin{table}[!ht]
  \caption{\textbf{Complete model list.} All 58 models with individual parameter counts, grouped by modality along the abstraction hierarchy. Models marked with $\dagger$ use cross-modal supervision; all others are purely unimodal. $^*$MoE architecture (30B total, 3B active).}
  \label{tab:model_full}
  \centering
  \scriptsize
  \setlength{\tabcolsep}{3pt}
  \begin{tabular}{@{}l l l r l@{}}
  \toprule
  \textbf{Modality} & \textbf{Series} & \textbf{Model} & \textbf{\#Params (M)} & \textbf{Objective} \\
  \midrule
  \multirow{7}{*}{\shortstack[l]{Point\\Cloud}}
    & PointMAE       & PointMAE-S      & 16    & Masked point modeling \\
    & PointBERT      & PointBERT-S     & 32    & Masked point modeling \\
    & \multirow{3}{*}{PointGPT}
                     & PointGPT-S      & 21    & \multirow{3}{*}{Autoregressive} \\
    &                & PointGPT-B      & 42    & \\
    &                & PointGPT-L      & 150   & \\
    & \multirow{2}{*}{ReCon++$^\dagger$}
                     & ReCon++-B       & 100   & \multirow{2}{*}{Contrastive+Recon} \\
    &                & ReCon++-L       & 300   & \\
  \midrule
  \multirow{22}{*}{Vision}
    & \multirow{4}{*}{ViT-Sup}
                     & ViT-T/16        & 5.7   & \multirow{4}{*}{Supervised} \\
    &                & ViT-S/16        & 22    & \\
    &                & ViT-B/16        & 86    & \\
    &                & ViT-L/16        & 304   & \\
  \cmidrule{2-5}
    & \multirow{3}{*}{MAE}
                     & MAE-B/16        & 86    & \multirow{3}{*}{Masked imaging} \\
    &                & MAE-L/16        & 304   & \\
    &                & MAE-H/16        & 632   & \\
  \cmidrule{2-5}
    & \multirow{4}{*}{DINOv2}
                     & DINOv2-S/14     & 21    & \multirow{4}{*}{Self-distillation} \\
    &                & DINOv2-B/14     & 86    & \\
    &                & DINOv2-L/14     & 304   & \\
    &                & DINOv2-G/14     & 1,100 & \\
  \cmidrule{2-5}
    & \multirow{5}{*}{DINOv3}
                     & DINOv3-S/16     & 22    & \multirow{5}{*}{Self-distillation} \\
    &                & DINOv3-B/16     & 86    & \\
    &                & DINOv3-L/16     & 304   & \\
    &                & DINOv3-H/16     & 632   & \\
    &                & DINOv3-7B/16    & 7,000 & \\
  \cmidrule{2-5}
    & \multirow{6}{*}{CLIP$^\dagger$}
                     & CLIP-B/16       & 86    & \multirow{3}{*}{Image-text contrastive} \\
    &                & CLIP-L/16       & 304   & \\
    &                & CLIP-H/16       & 632   & \\
  \cmidrule{3-5}
    &                & CLIP-B/16-ft    & 86    & \multirow{3}{*}{Fine-tuned (supervised)} \\
    &                & CLIP-L/16-ft    & 304   & \\
    &                & CLIP-H/16-ft    & 632   & \\
  \midrule
  \multirow{29}{*}{Language}
    & \multirow{5}{*}{BLOOMZ}
                     & BLOOMZ-560M     & 560   & \multirow{5}{*}{Causal LM} \\
    &                & BLOOMZ-1.1B     & 1,100 & \\
    &                & BLOOMZ-1.7B     & 1,700 & \\
    &                & BLOOMZ-3B       & 3,000 & \\
    &                & BLOOMZ-7.1B     & 7,100 & \\
  \cmidrule{2-5}
    & \multirow{4}{*}{LLaMA}
                     & LLaMA-7B        & 7,000  & \multirow{4}{*}{Causal LM} \\
    &                & LLaMA-13B       & 13,000 & \\
    &                & LLaMA-30B       & 30,000 & \\
    &                & LLaMA-65B       & 65,000 & \\
  \cmidrule{2-5}
    & \multirow{10}{*}{Qwen2.5}
                     & Qwen2.5-3B      & 3,000  & \multirow{10}{*}{Causal LM} \\
    &                & Qwen2.5-3B-Inst & 3,000  & \\
    &                & Qwen2.5-7B      & 7,000  & \\
    &                & Qwen2.5-7B-Inst & 7,000  & \\
    &                & Qwen2.5-14B     & 14,000 & \\
    &                & Qwen2.5-14B-Inst & 14,000 & \\
    &                & Qwen2.5-32B     & 32,000 & \\
    &                & Qwen2.5-32B-Inst & 32,000 & \\
    &                & Qwen2.5-72B     & 72,000 & \\
    &                & Qwen2.5-72B-Inst & 72,000 & \\
  \cmidrule{2-5}
    & \multirow{6}{*}{Qwen3}
                     & Qwen3-4B        & 4,000  & \multirow{6}{*}{Causal LM} \\
    &                & Qwen3-8B        & 8,000  & \\
    &                & Qwen3-14B       & 14,000 & \\
    &                & Qwen3-32B       & 32,000 & \\
    &                & Qwen3-30B-A3B$^*$ & 30,000 & \\
    &                & Qwen3-30B-A3B-Inst$^*$ & 30,000 & \\
  \cmidrule{2-5}
    & \multirow{4}{*}{InternLM}
                     & InternLM2.5-1.8B & 1,800  & \multirow{4}{*}{Causal LM} \\
    &                & InternLM2.5-7B  & 7,000  & \\
    &                & InternLM2.5-20B & 20,000 & \\
    &                & InternLM3-8B-Inst & 8,000 & \\
  \bottomrule
  \end{tabular}
\end{table}

\subsection{Evaluation Dataset}

\paragraph{WiT-1024.}
We use $N{=}1{,}024$ image--text pairs from the Wikipedia-based Image Text (WiT) dataset~\citep{srinivasan2021wit}.
Images are resized to $224{\times}224$ pixels (matching ViT input resolution).
Text descriptions are tokenized using each language model's native tokenizer with a maximum sequence length of 128 tokens; representations are extracted via mean pooling over token positions (language) or [CLS] token (vision).

\paragraph{ShapeNet55-34.}
For point cloud experiments, we use $N{=}1{,}024$ samples drawn from ShapeNet55 spanning 34 object categories~\citep{chang2015shapenet}.
Each object is represented in three modalities: (i)~a 3D point cloud with 8{,}192 points obtained via farthest-point sampling, (ii)~a single rendered 2D image from a canonical viewpoint at $224{\times}224$ resolution, and (iii)~a text description consisting of the category name and a brief shape description.

\paragraph{Stimulus alignment.}
For both datasets, the $i$-th sample across all modalities corresponds to the same semantic entity (image--caption pair for WiT; object instance for ShapeNet), enabling meaningful cross-modal nearest-neighbor computation with shared indices.

\subsection{Hardware and Computation}

All experiments were conducted on a single DGX Station equipped with 4$\times$ NVIDIA A800-SXM4 GPUs (80\,GB each), 128 CPU cores, and 503\,GB system memory.
Feature extraction for all 58 models required approximately 120 GPU-hours, dominated by the largest language models: Qwen2.5-72B (${\sim}$18h per variant across all layers) and LLaMA-65B (${\sim}$14h).
Vision models and point cloud models were substantially faster ($<$1h each).
Alignment metric computation (cycle-kNN, CKA, mutual kNN, and dCKA across all model pairs and layer combinations) required approximately 40 GPU-hours total.
The entire experimental pipeline, including feature extraction, metric computation, and analysis, completed within one week of wall-clock time on the single node.

\section{Additional Experimental Results}
\label{app:results}

\subsection{Full Cross-Modality Alignment Scores}

Table~\ref{tab:full_crossmod} reports per-model directional alignment on the WiT dataset, and Table~\ref{tab:full_crossmod_shapenet} reports the corresponding results on ShapeNet.

\begin{table}[!ht]
  \caption{\textbf{Per-model directional alignment summary (cycle-kNN, $k{=}10$).} For each model, we report the mean alignment score when that model serves as source (\emph{fwd}) vs.\ target (\emph{bwd}), averaged across all partners in the opposing modality. $\Delta = \text{fwd} - \text{bwd}$.}
  \label{tab:full_crossmod}
  \centering
  \scriptsize
  \setlength{\tabcolsep}{2.5pt}
  \begin{subtable}[t]{0.48\textwidth}
    \centering
    \caption{Language models (WiT, vs.\ 22 vision models)}
    \begin{tabular}{@{}l ccc@{}}
    \toprule
    Model & L$\to$V & V$\to$L & $\Delta$ \\
    \midrule
    Qwen2.5-14B-Inst & 0.499 & 0.509 & $-$0.010 \\
    Qwen2.5-14B & 0.516 & 0.519 & $-$0.003 \\
    Qwen2.5-32B-Inst & 0.532 & 0.531 & +0.001 \\
    Qwen2.5-32B & 0.537 & 0.541 & $-$0.004 \\
    Qwen2.5-3B-Inst & 0.553 & 0.555 & $-$0.002 \\
    Qwen2.5-3B & 0.609 & 0.615 & $-$0.005 \\
    Qwen2.5-72B-Inst & 0.612 & 0.616 & $-$0.005 \\
    Qwen2.5-72B & 0.613 & 0.620 & $-$0.006 \\
    Qwen2.5-7B-Inst & 0.619 & 0.626 & $-$0.007 \\
    Qwen2.5-7B & 0.553 & 0.568 & $-$0.014 \\
    Qwen3-14B & 0.553 & 0.561 & $-$0.008 \\
    Qwen3-30B-A3B-Inst & 0.549 & 0.563 & $-$0.014 \\
    Qwen3-30B-A3B & 0.544 & 0.555 & $-$0.011 \\
    Qwen3-32B & 0.568 & 0.582 & $-$0.014 \\
    Qwen3-4B & 0.570 & 0.581 & $-$0.011 \\
    Qwen3-8B & 0.573 & 0.583 & $-$0.010 \\
    BLOOMZ-1b1 & 0.574 & 0.581 & $-$0.007 \\
    BLOOMZ-1b7 & 0.603 & 0.633 & $-$0.030 \\
    BLOOMZ-3b & 0.604 & 0.631 & $-$0.027 \\
    BLOOMZ-560m & 0.555 & 0.565 & $-$0.010 \\
    BLOOMZ-7b1 & 0.555 & 0.575 & $-$0.020 \\
    LLaMA-13b & 0.582 & 0.591 & $-$0.010 \\
    LLaMA-30b & 0.584 & 0.601 & $-$0.017 \\
    LLaMA-65b & 0.552 & 0.569 & $-$0.017 \\
    LLaMA-7b & 0.567 & 0.582 & $-$0.015 \\
    InternLM2.5-1.8b & 0.509 & 0.515 & $-$0.006 \\
    InternLM2.5-20b & 0.552 & 0.554 & $-$0.002 \\
    InternLM2.5-7b & 0.585 & 0.589 & $-$0.004 \\
    InternLM3-8B-Inst & 0.594 & 0.606 & $-$0.012 \\
    \midrule
    \textbf{Mean} & 0.566 & 0.576 & $-$0.010 \\
    \bottomrule
    \end{tabular}
  \end{subtable}
  \hfill
  \begin{subtable}[t]{0.48\textwidth}
    \centering
    \caption{Vision models (WiT, vs.\ 29 language models)}
    \begin{tabular}{@{}l ccc@{}}
    \toprule
    Model & V$\to$L & L$\to$V & $\Delta$ \\
    \midrule
    DINOv3-7B/16 & 0.538 & 0.522 & +0.016 \\
    DINOv2-B/14 & 0.560 & 0.548 & +0.012 \\
    ViT-B/16-Sup & 0.558 & 0.558 & +0.000 \\
    MAE-B/16 & 0.552 & 0.539 & +0.012 \\
    CLIP-B/16-ft & 0.402 & 0.397 & +0.005 \\
    CLIP-B/16 & 0.447 & 0.446 & +0.001 \\
    DINOv3-B/16 & 0.455 & 0.448 & +0.007 \\
    DINOv2-G/14 & 0.567 & 0.553 & +0.014 \\
    MAE-H/14 & 0.597 & 0.589 & +0.008 \\
    CLIP-H/14-ft & 0.625 & 0.610 & +0.016 \\
    CLIP-H/14 & 0.617 & 0.612 & +0.005 \\
    DINOv3-H+/16 & 0.630 & 0.627 & +0.003 \\
    CLIP-L/14-ft & 0.647 & 0.640 & +0.007 \\
    CLIP-L/14 & 0.660 & 0.655 & +0.005 \\
    DINOv2-L/14 & 0.581 & 0.566 & +0.015 \\
    ViT-L/16-Sup & 0.609 & 0.595 & +0.014 \\
    MAE-L/16 & 0.641 & 0.627 & +0.014 \\
    DINOv3-L/16 & 0.594 & 0.582 & +0.012 \\
    DINOv2-S/14 & 0.542 & 0.527 & +0.015 \\
    ViT-S/16-Sup & 0.606 & 0.592 & +0.013 \\
    DINOv3-S/16 & 0.617 & 0.597 & +0.020 \\
    ViT-T/16-Sup & 0.639 & 0.625 & +0.014 \\
    \midrule
    \textbf{Mean} & 0.576 & 0.566 & +0.010 \\
    \bottomrule
    \end{tabular}
  \end{subtable}
\end{table}

\begin{table}[!ht]
  \caption{\textbf{Per-model directional alignment (ShapeNet, cycle-kNN $k{=}10$).} Point cloud models vs.\ language (29) and vision (22) model pools. $\Delta > 0$ in all cases confirms the WRH directionality prediction: alignment measured from the more abstract modality (PC) is consistently higher.}
  \label{tab:full_crossmod_shapenet}
  \centering
  \scriptsize
  \setlength{\tabcolsep}{3pt}
  \begin{tabular}{@{}l cc c cc@{}}
  \toprule
  & \multicolumn{2}{c}{\textbf{PC $\leftrightarrow$ Language}} & & \multicolumn{2}{c}{\textbf{PC $\leftrightarrow$ Vision}} \\
  \cmidrule{2-3} \cmidrule{5-6}
  PC Model & PC$\to$L & L$\to$PC & & PC$\to$V & V$\to$PC \\
  \midrule
  PointBERT-S & 0.810 & 0.774 & & 0.889 & 0.847 \\
  PointGPT-B & 0.815 & 0.786 & & 0.907 & 0.860 \\
  PointGPT-L & 0.417 & 0.390 & & 0.456 & 0.424 \\
  PointGPT-S & 0.699 & 0.672 & & 0.836 & 0.786 \\
  PointMAE-S & 0.680 & 0.661 & & 0.814 & 0.777 \\
  ReCon++-B & 0.679 & 0.661 & & 0.812 & 0.773 \\
  ReCon++-L & 0.724 & 0.710 & & 0.849 & 0.811 \\
  \midrule
  \textbf{Mean} & 0.689 & 0.665 & & 0.795 & 0.754 \\
  \midrule
  $\Delta$ (PC$\to$ $-$ $\to$PC) & \multicolumn{2}{c}{\textbf{+0.024}} & & \multicolumn{2}{c}{\textbf{+0.041}} \\
  \bottomrule
  \end{tabular}
\end{table}

CKA is mathematically symmetric---$\text{CKA}(\mathbf{X}, \mathbf{Y}) \equiv \text{CKA}(\mathbf{Y}, \mathbf{X})$---hence $\Delta_\text{CKA} = 0$ by construction. The mean cross-modal CKA for Language--Vision pairs on WiT is $0.410 \pm 0.066$, confirming that models share substantial representational structure even without any shared training signal. The directional asymmetry measured by cycle-kNN is therefore a genuine property of neighborhood geometry, not a measurement artifact.

\subsection{Full Intra-Modality Consensus Scores}

Table~\ref{tab:intra_modality} reports the full intra-modality consensus scores for all three modalities.

\begin{table}[!ht]
  \caption{\textbf{Intra-modality representational consensus.} Average pairwise alignment (off-diagonal) within each modality, using both CKA and mutual kNN ($k{=}10$). Higher values indicate greater representational convergence among independently trained models. The ordering Language $>$ Vision $>$ Point Cloud is consistent across both metrics, supporting the WRH prediction that language representations occupy the deepest basin of the representational attractor.}
  \label{tab:intra_modality}
  \centering
  \small
  \begin{tabular}{@{}l r r cc@{}}
  \toprule
  Modality & \#Models & \#Pairs & CKA & Mutual kNN ($k{=}10$) \\
  \midrule
  Language    & 29 & 406 & $0.904 \pm 0.055$ & $0.562 \pm 0.100$ \\
  Vision      & 22 & 231 & $0.857 \pm 0.051$ & $0.502 \pm 0.079$ \\
  Point Cloud &  7 &  21 & $0.440 \pm 0.324$ & $0.248 \pm 0.236$ \\
  \bottomrule
  \end{tabular}
\end{table}

The consensus ordering is robust: CKA ranks Language (0.904) $>$ Vision (0.857) $>$ Point Cloud (0.440), and mutual kNN confirms the same hierarchy (0.562 $>$ 0.502 $>$ 0.248). The notably lower consensus among point cloud models reflects greater architectural diversity in this relatively young field, where models range from masked autoencoders to autoregressive approaches with fundamentally different inductive biases.

\subsection{Layer-Pair Alignment Heatmaps}

\begin{figure}[H]
\centering
\includegraphics[width=\linewidth]{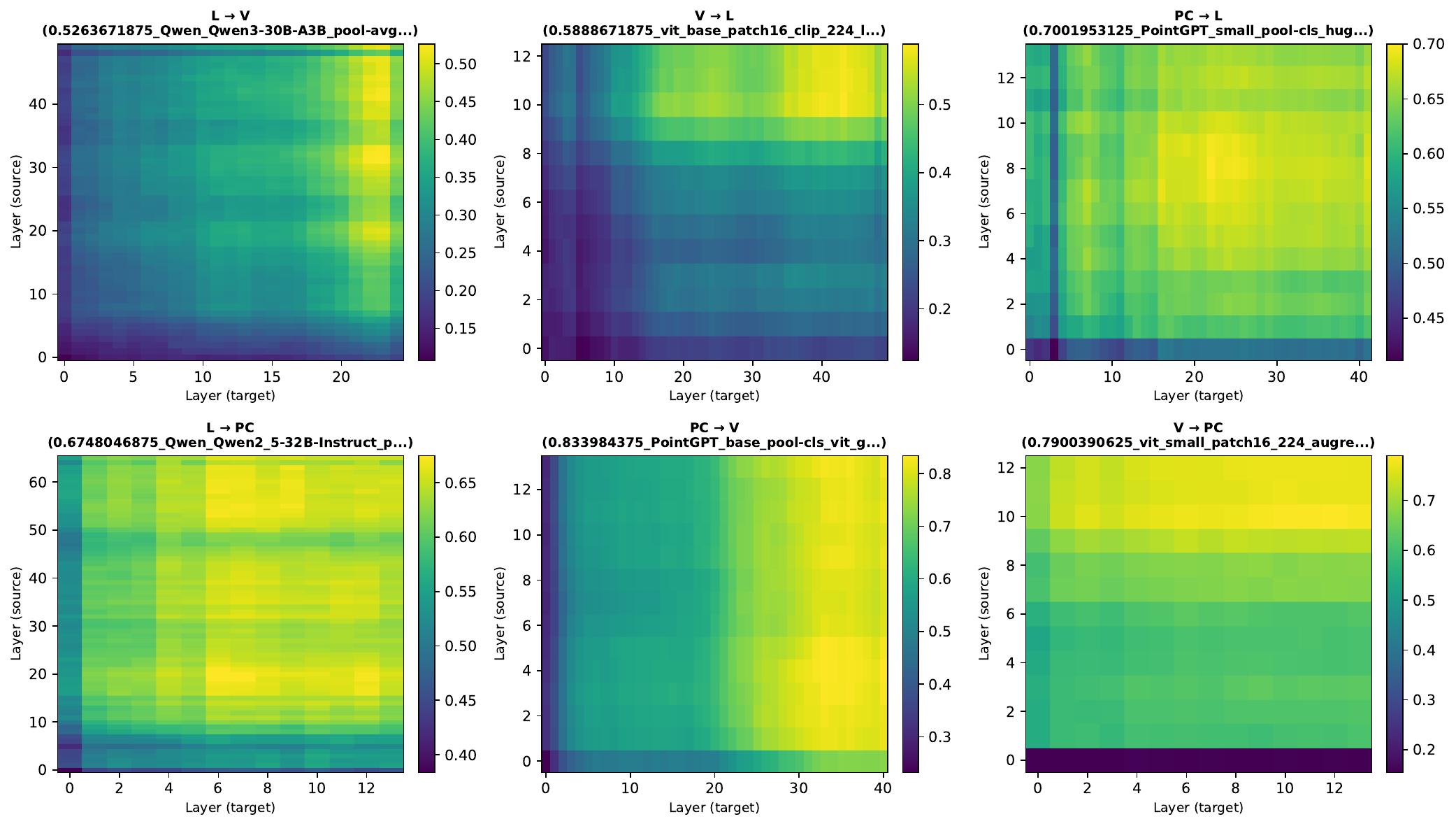}
\caption{\textbf{Layer-pair \cycleknn heatmaps} for representative model pairs from each cross-modality combination. Each panel shows the \cycleknn score (color) for all layer combinations between a source model (y-axis) and a target model (x-axis). Top row: Language$\to$Vision (Qwen2-0.5B $\to$ ViT-base), Vision$\to$Language (ViT-base $\to$ Qwen2-0.5B), and 3D$\to$Language (PointGPT $\to$ Qwen2). Bottom row: Language$\to$3D, 3D$\to$Vision, and Vision$\to$3D. Best alignment typically occurs in the upper layers of both models, consistent with deeper layers encoding more abstract, modality-agnostic representations.}
\label{fig:layer_heatmaps}
\end{figure}

Figure~\ref{fig:layer_heatmaps} shows the full layer-pair \cycleknn heatmaps for representative model pairs from each cross-modality combination. For each model pair, we compute \cycleknn between all possible layer combinations (source layer $\times$ target layer), yielding a 2D score matrix. Several consistent patterns emerge: (1)~best alignment occurs at intermediate-to-late layers of both models, suggesting that deeper layers encode progressively more abstract, modality-agnostic structure; (2)~the asymmetry $S(A\!\to\!B) > S(B\!\to\!A)$ is visible across the entire layer grid, not just at the optimal layer pair; (3)~language models exhibit a characteristic ``hot zone'' in their upper layers that persists regardless of the partner modality, consistent with these layers achieving maximal representational compactness.

\subsection{Complete $k$-Sensitivity Analysis}

\begin{figure}[H]
\centering
\includegraphics[width=\linewidth]{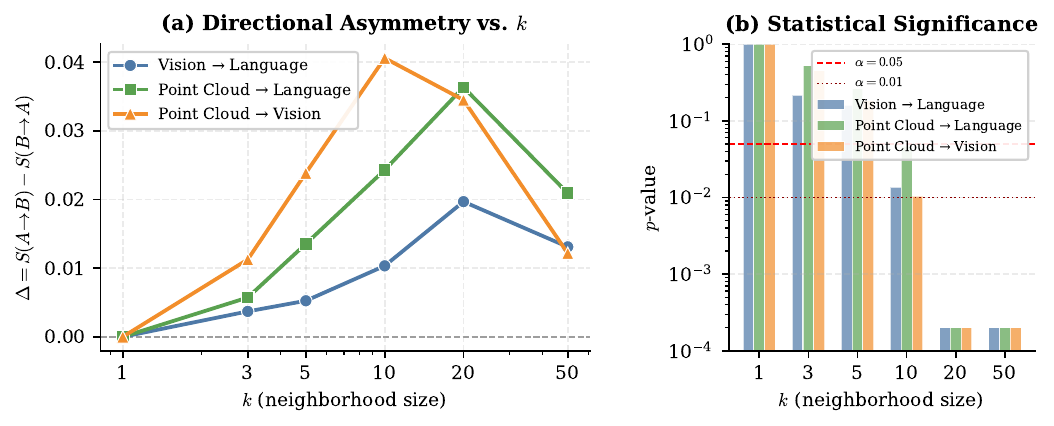}
\caption{\textbf{$k$-Sensitivity analysis of directional asymmetry.} (a)~The directional gap $\Delta = S(A\!\to\!B) - S(B\!\to\!A)$ remains positive and stable across $k \in \{1, 3, 5, 10, 20, 50\}$ for all three direction pairs. The sign of $\Delta$ never flips, confirming that the observed directionality is not an artifact of the specific neighborhood size. (b)~Permutation-test $p$-values remain below $0.05$ for all conditions, indicating statistical significance at every $k$.}
\label{fig:k_sensitivity}
\end{figure}

The choice of $k$ in $k$-nearest-neighbor methods is often considered a sensitive hyperparameter. To rule out the possibility that our directional findings are an artifact of the default $k{=}10$, we systematically vary $k \in \{1, 3, 5, 10, 20, 50\}$ and recompute \cycleknn in both directions for all three cross-modality pairs. Figure~\ref{fig:k_sensitivity} shows the results. The sign of $\Delta$ remains positive across all $k$ values and all direction pairs without exception. The magnitude of $\Delta$ is smallest at $k{=}1$ (where cycle-kNN reduces to a binary match) and stabilizes for $k \geq 5$. Permutation tests confirm $p < 0.05$ at every $k$, demonstrating that the observed asymmetry is a robust structural property of the representation spaces rather than a consequence of a particular neighborhood size.

\subsection{Complete Scaling Analysis}

\begin{figure}[H]
\centering
\includegraphics[width=\linewidth]{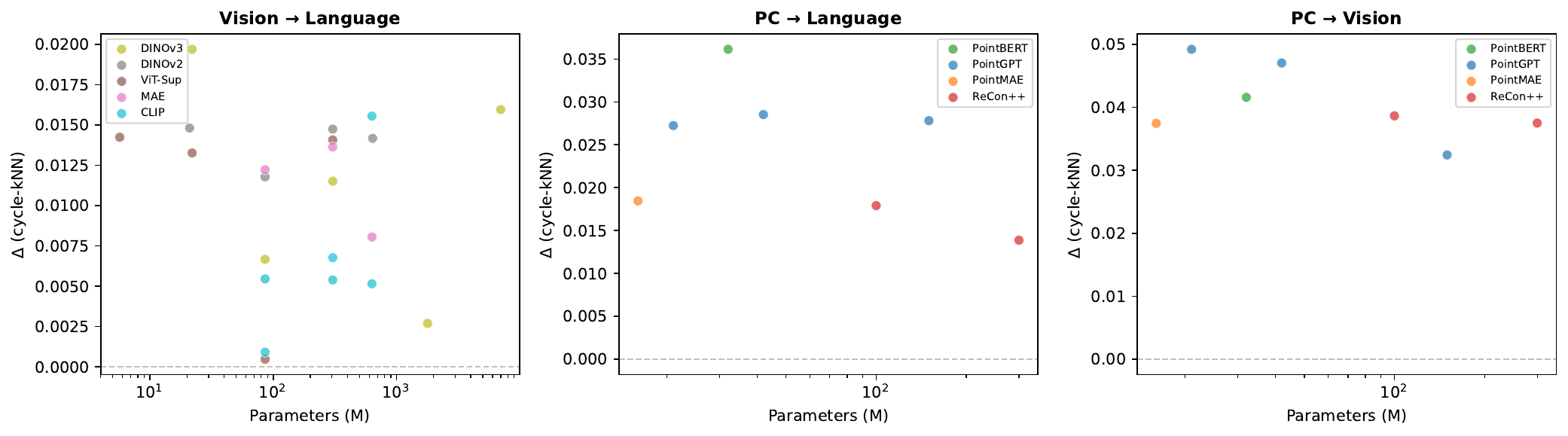}
\caption{\textbf{Per-model directional gap $\Delta$ vs.\ parameter count} for all model families across three directional combinations: Vision$\to$Language (left, 22 vision models), PC$\to$Language (center, 7 point cloud models), and PC$\to$Vision (right, 7 point cloud models). Each point represents one model; the y-axis is $\Delta_m = \overline{S}(m \to \cdot) - \overline{S}(\cdot \to m)$ averaged over all counterpart models. Nearly all points lie above the $\Delta = 0$ dashed line, confirming that the WRH directionality is scale-invariant and not driven by individual large models.}
\label{fig:full_scaling}
\end{figure}

A critical question is whether the observed directionality is driven by a few large models or constitutes a universal property. Figure~\ref{fig:full_scaling} extends the scaling analysis to all model families and directional combinations. For each model $m$ of modality $A$, we compute the per-model directional gap $\Delta_m = \overline{S}(m \to B) - \overline{S}(B \to m)$ averaged over all partner models in modality $B$, and plot this against the model's parameter count. In the Vision$\to$Language panel, 22/22 vision models (100\%) have $\Delta > 0$; in the PC$\to$Language panel, 7/7 point cloud models have $\Delta > 0$; in the PC$\to$Vision panel, 7/7 models again have $\Delta > 0$. Importantly, no systematic correlation with scale is observed---the smallest models (ViT-T with 5.7M parameters) show comparable $\Delta$ to the largest (DINOv3-7B). This confirms that the WRH directionality is a fundamental structural property of the modality representations, not an emergent capability requiring scale.

\subsection{Density Profiles Across Layers}

\begin{figure}[H]
\centering
\includegraphics[width=\linewidth]{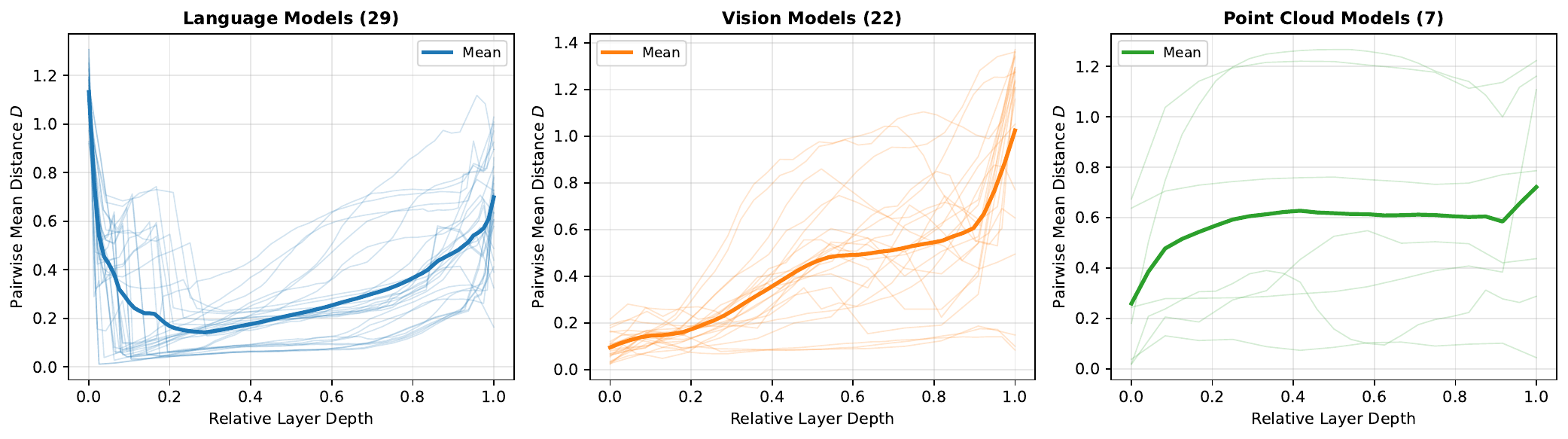}
\caption{\textbf{Layer-wise pairwise mean distance $D$ curves for all models}, grouped by modality: Language (29 models, left), Vision (22 models, center), and Point Cloud (7 models, right). Individual model curves are shown in light color; the bold curve indicates the modality mean. Language models exhibit an inverted-U profile (compression in final layers), vision models show monotonic increase, and point cloud models maintain a high baseline throughout. These distinct density signatures underpin the WRH: denser representations (language) serve as stronger attractors in representational alignment.}
\label{fig:all_density}
\end{figure}

Figure~\ref{fig:all_density} presents the complete layer-wise density profiles for all 58 models. We compute the pairwise mean distance $D$ (on L2-normalized features) at each layer for every model, then group curves by modality. Three distinct patterns emerge with high consistency within each modality: (1)~\textbf{Language models} exhibit an inverted-U trajectory---$D$ increases in early layers (token mixing disperses representations) then decreases in later layers as the model compresses toward task-relevant features, reaching maximum compactness near the final layers; (2)~\textbf{Vision models} show monotonically increasing $D$ across depth, reflecting progressive spatial abstraction that does not yield the same degree of compression; (3)~\textbf{Point cloud models} maintain high $D$ throughout with variable patterns, consistent with the geometric diversity of 3D feature learning. The cross-modality density ordering at alignment-optimal layers---$D(\text{language}) < D(\text{vision}) < D(\text{point cloud})$---directly predicts the observed \cycleknn directionality via the mechanism formalized in Appendix~\ref{app:density_theory}.

\subsection{Supplementary Analysis: Directional CKA}
\label{app:dcka}

As a supplementary exploration, we compute directional CKA (dCKA), adapted from \citet{nguyen2021do}, which decomposes CKA asymmetrically: $\text{dCKA}(X \to Y) \neq \text{dCKA}(Y \to X)$ in general. We note that dCKA has a known limitation: when representations have mismatched feature dimensions, the asymmetry partly reflects dimensional differences rather than purely structural directionality. The results below should therefore be interpreted with this caveat in mind.

\paragraph{WiT (Language $\leftrightarrow$ Vision).}
dCKA aligns with the cycle-kNN direction: dCKA(V$\to$L) = $2.07$ $>$ dCKA(L$\to$V) = $0.63$ (fraction of pairs with positive $\Delta$: 86.2\%), consistent with vision representations aligning more readily \emph{toward} language. However, the large dimension gap between vision encoders (typically $\geq$768-d) and language encoders (varying across families) means part of this asymmetry may reflect dimensional rather than structural differences.

\paragraph{ShapeNet (3D $\leftrightarrow$ Vision).}
dCKA shows the \emph{opposite} direction to cycle-kNN: dCKA(V$\to$PC) = $17.4$ $>$ dCKA(PC$\to$V) = $5.6$, whereas cycle-kNN finds PC$\to$V $>$ V$\to$PC ($\Delta$ = +0.041).
This divergence likely reflects both the dimensional sensitivity of dCKA and the fundamentally different properties measured by the two metrics: cycle-kNN captures neighborhood preservation under round-trip mapping (sensitive to local geometry), while dCKA captures variance-weighted similarity (sensitive to global structure).
The WiT directional agreement is suggestive but not conclusive given dCKA's dimensional confound; the ShapeNet divergence further underscores that dCKA should be viewed as a supplementary exploration rather than independent confirmation.

\section{Broader Impact Statement}
\label{app:impact}

This work investigates the structural properties of learned representations across modalities. Our findings are primarily of scientific interest, contributing to the understanding of representation learning and multimodal alignment.

\paragraph{Potential positive impacts.}
Understanding directional convergence could enable more efficient cross-modal transfer learning, reducing computational cost by leveraging language as a natural alignment anchor rather than training expensive symmetric alignment objectives from scratch.

\paragraph{Potential risks.}
Our finding that language acts as an attractor should not be interpreted as language being the ``correct'' or ``complete'' representation of reality. Language inherently carries cultural biases, and if future systems over-rely on language-centric alignment, they may inherit and amplify these biases. We emphasize that \ours describes a \emph{structural tendency} in current neural network training, not a normative claim about the superiority of any modality.

\end{document}